\documentclass[letterpaper]{article}
\usepackage[preprint]{aaai2027}
\usepackage[hyphens]{url}
\usepackage{graphicx}
\usepackage{natbib}
\usepackage{caption}
\usepackage{algorithm}
\usepackage{algorithmic}
\usepackage{booktabs}
\usepackage{tabularx}
\usepackage{array}
\usepackage{amsmath}
\nocopyright

\title{Training-Free Knowledge Transfer Across Model Scales through Activation-Guided Pruning}
\author{
    Jiahe Fan\textsuperscript{1},
    Si Chen\textsuperscript{2},
    Yinghao Hou\textsuperscript{1},
    Aiyuan Zhang\textsuperscript{1},
    Hong Xie\textsuperscript{3}
}
\affiliations{
    \textsuperscript{1}University of Science and Technology of China\\
    \textsuperscript{2}School of Information Science and Technology, Department of Automation,
    University of Science and Technology of China\\
    \textsuperscript{3}School of Computer Science and Technology, University of Science and Technology of China\\
    fanjiahe@mail.ustc.edu.cn
}

\begin{document}
\raggedbottom

\maketitle

\begin{abstract}
Heterogeneous model fusion combines models that differ in tasks, initializations, architectures, or scales.
We study an underexplored cross-scale setting: improving a small recipient language model with a stronger donor despite substantial architectural mismatch.
We ask whether useful capabilities can be transferred without explicit neuron-wise semantic alignment.
Building on the observation that truncating a large model to a smaller architecture and injecting it with a tiny mixing weight can improve the recipient, we propose Activation-Prune-Merge (APM), an activation-guided framework for cross-scale fusion.
APM constructs task-conditioned activation maps on the donor, selects salient layers, hidden dimensions, attention heads, and MLP neurons to prune it to the recipient architecture, and injects the resulting donor slice into the original recipient using a micro interpolation coefficient.
This formulation treats the donor as a source of concentrated functional components rather than requiring precise structural transplantation.
Across 16 benchmarks spanning reasoning, mathematics, code generation, instruction following, and classification, APM improves the overall average accuracy from 55.5\% to 60.6\% over the original 3B recipient.
RTE accuracy increases from 64.3\% to 82.3\%, QNLI from 52.3\% to 65.7\%, and BoolQ from 70.8\% to 79.2\%.
Analyses of injection ratios and sequential multi-stage fusion further show that activation-guided extraction improves the quality of the transferable donor slice while preserving the small-ratio fusion regime.
These results show that cross-scale heterogeneous fusion succeeds without explicit semantic alignment when the donor contribution is sufficiently concentrated and carefully selected.
\end{abstract}

\begin{figure}[t]
\centering
\includegraphics[width=0.95\columnwidth]{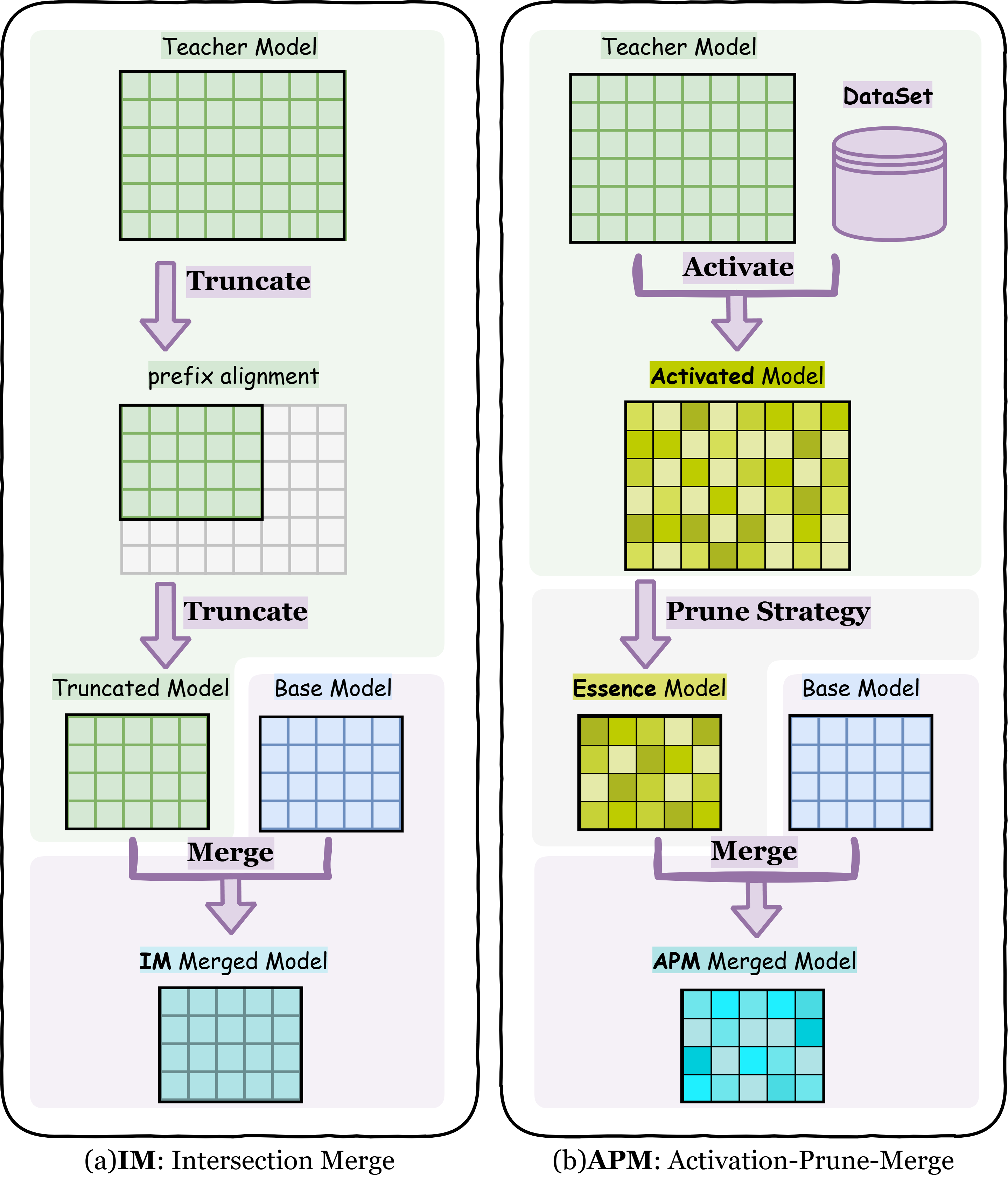}
\caption{Intersection-Merge (IM) versus activation-guided cross-scale fusion.}
\label{fig:pipeline}
\end{figure}

\section{Introduction}

Model merging combines the capabilities of multiple trained models within a single model and has proved effective for consolidating fine-tuned checkpoints, composing task-specific skills, and building multi-task or domain-specialized models \citep{wortsman2022soups,ilharco2023task,yadav2023ties,yu2024language}.

Classical homogeneous merging methods combine checkpoints or task vectors through parameter averaging, vector arithmetic, conflict resolution, or sparsification \citep{wortsman2022soups,matena2022merging,ilharco2023task,yadav2023ties,yu2024language}.
However, they require the models to share the same architecture and compatible parameterization, which substantially restricts the set of models that can be merged directly.
Approaches to heterogeneous transfer include knowledge distillation and adapter- or projection-based transformations, while alignment-based methods establish feature or semantic correspondences before fusion \citep{hinton2015distilling,pfeiffer2021adapterfusion,gu2025seme,stoica2024zipit}.
Yet many of these methods require additional training, learned transformations, optimization, or explicit cross-model alignment, making heterogeneous fusion considerably more involved.

Prior work introduced Intersection-Merge (IM), which directly truncates a large model to a target architecture by retaining the leading parameter indices and injects the resulting donor slice into a smaller model with a tiny mixing weight \citep{fan2026rethinkingheterogeneousllmmerging}.
Large language models at different scales exhibit a surprising form of transferability: a stronger donor can improve a smaller recipient without neuron-level semantic alignment.
Under this view, IM supplies a coarsely selected donor slice.
Building on this baseline, we ask a sharper question: can we extract a more task-relevant donor slice so that the same micro-injection budget produces larger gains?

We propose Activation-Prune-Merge (APM), a training-free framework that extracts task-relevant structure from a strong donor without explicit neuron-wise semantic alignment (Figure~\ref{fig:pipeline}).
Given task data, APM first runs the donor model and collects activation statistics over a specified token range.
It then uses the resulting activation maps to select salient layers, hidden dimensions, attention heads, and MLP neurons.
These components are pruned and arranged into the exact architecture of the recipient, after which the resulting donor slice is injected into the original recipient with a micro interpolation coefficient.
APM concentrates the donor contribution before fusion while preserving the recipient as the dominant component of the merged model.

Experiments across reasoning, mathematics, code generation, instruction following, and classification tasks show that APM improves both the original 3B recipient and IM.
The gains are especially pronounced on natural language inference and reasoning benchmarks.
At small injection ratios, cross-scale heterogeneous fusion behaves less like neuron-wise transplantation, which would require strict semantic alignment, and more like adding a concentrated donor extract to the recipient.
Additional analyses of injection ratios and sequential multi-stage fusion indicate that activation-guided extraction remains effective under small merge coefficients and that an APM-fused model can continue to benefit from a subsequent donor injection.
Together, these findings support concentration transfer as a useful perspective on cross-scale heterogeneous fusion.

Our contributions are threefold:
\begin{enumerate}
\item \textbf{APM framework.} We introduce APM, an activation-guided heterogeneous fusion pipeline that combines donor activation mapping, target-shape pruning, and micro-injection into a smaller recipient model. Under the same micro-injection rule, activation-guided pruning improves cross-scale transfer beyond IM, indicating that APM extracts a more effective and task-relevant donor slice.
\item \textbf{Transfer view.} Our results support a concentration-transfer view: small-ratio heterogeneous fusion depends on concentrated task-relevant donor information rather than strict neuron-wise semantic correspondence. This explains why activation-guided selection can outperform coarse truncation under the same micro-injection budget.
\item \textbf{Broad evaluation.} We evaluate APM across benchmark families, injection ratios, and sequential fusion settings, covering mathematical reasoning, code generation, instruction following, language understanding, commonsense knowledge, and multi-task reasoning. The results show gains beyond a single benchmark or fusion configuration, including under repeated small-ratio injection.
\end{enumerate}

\section{Related Work}

\subsection{Homogeneous Model Merging}

Homogeneous model merging combines parameters or task updates from models with a shared architecture \citep{lu2024merge,song2026model,yang2026model}, supported geometrically by low-loss paths and linear mode connectivity between compatible solutions \citep{garipov2018loss,frankle2020linear}.
Checkpoint averaging and Fisher-weighted merging combine parameters directly, whereas Task Arithmetic, TIES-Merging, DARE, Model Breadcrumbs, and task localization operate on task vectors through arithmetic, trimming, sign resolution, sparsification, or selective updates \citep{wortsman2022soups,matena2022merging,ilharco2023task,yadav2023ties,yu2024language,davari2024breadcrumbs,he2024localize}.
Complementary work studies dataless knowledge fusion, task-subspace matching, and task-information localization \citep{jin2023dataless,tam2024merging,wang2024localizing}, while benchmarks and scaling analyses examine specialized LLM merging, model selection, and performance as merge count or model scale grows \citep{he2025mergebench,yadav2024what,wang2025mergingscaling,hitit2026systematic}.
Training-free alignment methods such as Git Re-Basin and REPAIR establish parameter correspondences or correct activation statistics before merging structurally compatible models \citep{ainsworth2023git,jordan2023repair}.

\subsection{Heterogeneous and Cross-Scale Model Fusion}

Heterogeneous fusion spans models with different architectures or scales \citep{chen2026heterofusion,soro2026lsmerge}; HM3 treats label-space heterogeneity by merging same-architecture classifiers with distinct outputs \citep{hackmann2024hm3}.
Knowledge distillation transfers behavior by training a recipient on a donor's output distribution without combining mismatched parameters \citep{hinton2015distilling}, whereas adapter-based methods learn task-specific intermediates and compose them through an additional fusion mechanism \citep{pfeiffer2021adapterfusion}.
Alignment-based methods establish semantic, feature, or structural correspondences before fusion \citep{gu2025seme}. ZipIt! extends training-free fusion across tasks by matching and merging intermediate features while accommodating limited architectural differences \citep{stoica2024zipit}.
Other methods package knowledge into modular skills, fuse chat-model capabilities, or optimize merge coefficients for heterogeneous multimodal models \citep{du2025graftllm,wan2024knowledgellm,wan2024fusechat,du2025adamms}.
Many such approaches require training, optimization, learned intermediates, or explicit correspondence; APM instead uses activations to rank donor-internal components, without training or cross-model feature or neuron-wise semantic alignment.

\subsection{Intersection-Merge}

\citet{fan2026rethinkingheterogeneousllmmerging} proposed Intersection-Merge (IM), which retains leading donor layers and parameter indices until the donor matches the recipient shape, then applies a very small linear interpolation coefficient.
IM showed that a target-shaped donor slice can transfer capability without training or explicit semantic alignment.
Its static front-aligned extraction, however, is not conditioned on the evaluated task.
APM retains the same cross-scale injection regime while replacing this rule with task-conditioned donor selection.

\subsection{Activation-Guided Structured Pruning}

Activation-based importance estimation is widely used for structured pruning.
Minitron scores layers, heads, hidden dimensions, and MLP neurons from calibration activations, while SliceGPT and Wanda provide complementary representation- or activation-aware pruning strategies \citep{muralidharan2024minitron,ashkboos2024slicegpt,sun2023wanda}.
These methods optimize a standalone compressed model.
APM instead profiles a task-conditioned donor, extracts a recipient-shaped slice, and micro-injects it into an independently trained recipient.
Its novelty lies in using activation saliency for transfer-oriented donor selection within training-free cross-scale fusion.

\section{Method}

\begin{figure*}[t]
\centering
\includegraphics[width=\textwidth]{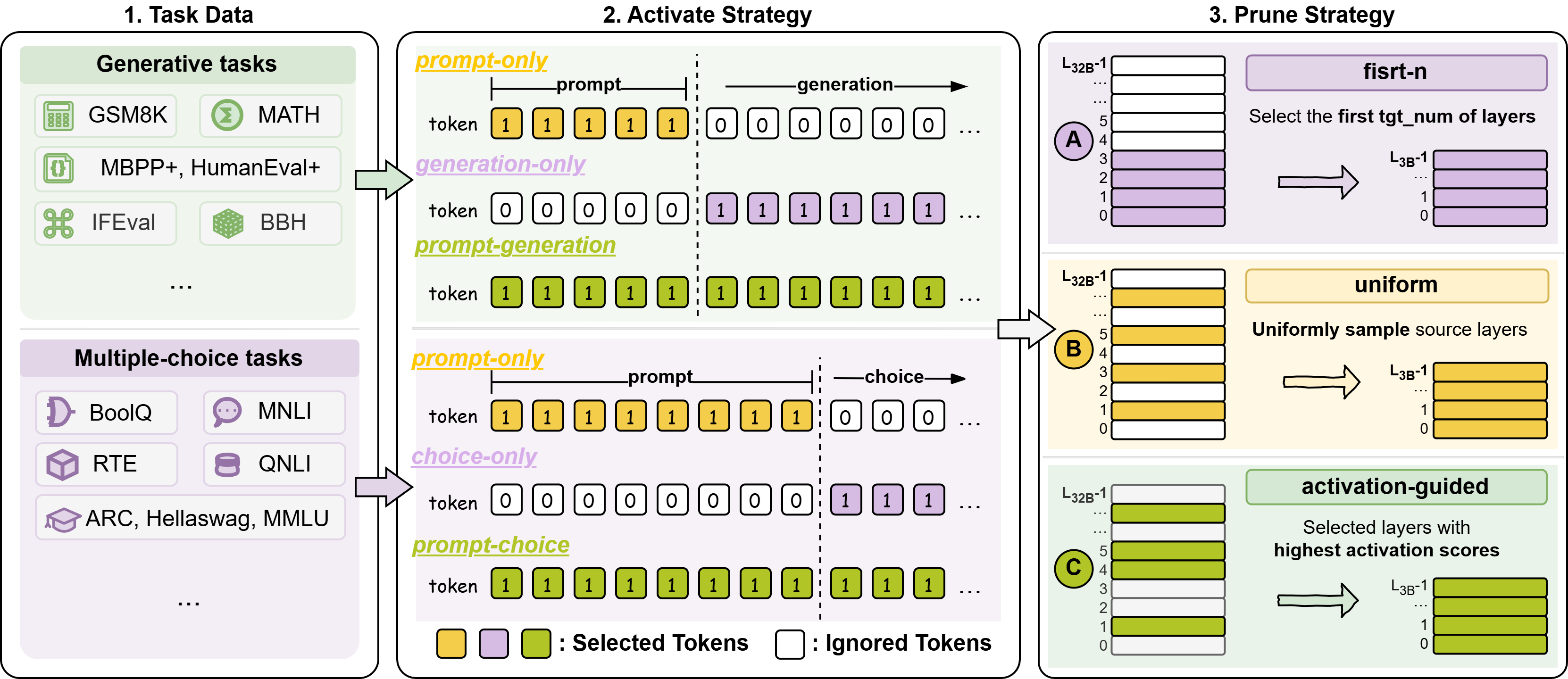}
\caption{Overview of task-conditioned activation profiling and layer selection in APM.
Task examples are divided into generative and multiple-choice settings.
The activation strategy specifies which prompt, generation, or answer-choice tokens contribute to the activation statistics.
The resulting scores support three target-depth pruning strategies: first-$n$, uniform, and activation-guided layer selection.}
\label{fig:activation-pruning-overview}
\end{figure*}

\subsection{Preliminaries}

\subsubsection{Problem Formulation}

Let $M_s=(\mathcal{A}_s,\theta_s)$ be a small recipient model and $M_l=(\mathcal{A}_l,\theta_l)$ a larger donor from the same model family.
Their depths, widths, and attention structures differ.
Given task data $\mathcal{D}$, we seek a fused model that retains the architecture and inference cost of $M_s$ while absorbing task-relevant donor components.
Formally, we seek a training-free extraction operator
$\mathcal{P}_{\mathcal{D}}:\theta_l\mapsto\tilde{\theta}_l$ such that
$\tilde{\theta}_l$ has architecture $\mathcal{A}_s$ and can be fused directly
with $\theta_s$. APM instantiates this operator through task-conditioned
activation profiling and target-shape pruning, followed by micro-injection.
Figure~\ref{fig:activation-pruning-overview} illustrates the extraction stage.

\subsubsection{Intersection-Merge}

Intersection-Merge (IM) is the direct starting point of APM
\citep{fan2026rethinkingheterogeneousllmmerging}. Let
$\mathcal{I}^{\mathrm{front}}(\mathcal{A}_l,\mathcal{A}_s)$ contain the leading
donor layers and the leading indices of every tensor dimension required by the
recipient. IM constructs a target-shaped donor by
\begin{equation}
\begin{gathered}
\theta_l^{\mathrm{IM}}
=\mathcal{P}_{\mathrm{front}}
\!\left(\theta_l;
\mathcal{I}^{\mathrm{front}}(\mathcal{A}_l,\mathcal{A}_s)\right),\\
\operatorname{shape}(\theta_l^{\mathrm{IM}})
=\operatorname{shape}(\theta_s).
\end{gathered}
\label{eq:im-extraction}
\end{equation}
and injects it with a small interpolation coefficient:
\begin{equation}
\theta_{\mathrm{IM}}
=(1-\mu)\theta_s+\mu\theta_l^{\mathrm{IM}},
\qquad 0<\mu\ll 1.
\label{eq:im-merge}
\end{equation}
Thus, IM resolves cross-scale tensor mismatch without training or semantic
alignment, but its donor selection is fixed and task-agnostic. APM retains
Equation~\ref{eq:im-merge} as the controlled fusion rule and replaces
the fixed front-aligned operator $\mathcal{P}_{\mathrm{front}}$ with the task-conditioned operator
$\mathcal{P}_{\mathcal{D}}$. This isolates the effect of selecting a more
informative donor slice.

\subsection{Task-Conditioned Activation Mapping}

We run the donor on $\mathcal{D}$ and accumulate activation magnitudes for four structural units: transformer layers, hidden channels, attention heads, and MLP neurons.
Only task-relevant token positions contribute to these statistics.
For generative tasks, the profiling scope can cover prompt tokens, generation tokens, or both; for multiple-choice tasks, it can cover prompt tokens, answer choices, or both.
Let $\Omega(x)$ be the profiled token positions for input $x$, and let $Z=\sum_{x\in\mathcal{D}}|\Omega(x)|$.
For structural unit $u$ with activation $g_u(x,t)$, its importance score is
\begin{equation}
s_u=\frac{1}{Z}\sum_{x\in\mathcal{D}}\sum_{t\in\Omega(x)}
\phi_u\!\left(g_u(x,t)\right),
\label{eq:activation-score}
\end{equation}
where $\phi_u(\cdot)$ is the $\ell_2$ norm for layer and attention-head vectors and the absolute value for hidden-channel and MLP-neuron scalars.
The resulting scores form a task-conditioned concentration map used to rank donor components.
They are accumulated online without storing full activation tensors; the supplementary material gives the unit-specific expansions of Equation~\ref{eq:activation-score}.

\subsection{Activation-Guided Target-Shape Pruning}

APM converts the activation map into a dense donor slice with exactly the
recipient's depth, residual width, attention configuration, and MLP width. Let
these target cardinalities be $L_s$, $d_s$, $(H_s,H_s^{\mathrm{KV}})$, and
$m_s$, respectively. For depth, the implementation supports three controlled
strategies:
\begin{equation}
\mathcal{I}_{L}=
\begin{cases}
\{0,\ldots,L_s-1\}, & \text{first},\\
\operatorname{Uniform}(L_l,L_s), & \text{uniform},\\
\operatorname{TopK}(S_{\mathrm{layer}},L_s), & \text{activation}.
\end{cases}
\label{eq:layer-selection}
\end{equation}
All selected layer indices are sorted to preserve donor-depth order. The other
dimensions are selected by activation score:
\begin{equation}
\mathcal{I}_{d}
=\operatorname{TopK}(S_{\mathrm{hidden}},d_s),\qquad
\mathcal{I}_{m}^{(\ell)}
=\operatorname{TopK}(S_{\mathrm{mlp}}^{(\ell)},m_s).
\label{eq:hidden-selection}
\end{equation}
The hidden-channel set is global, whereas MLP neurons are selected separately
for every retained layer $\ell$. For grouped-query attention, query and
key--value (KV) heads must be treated separately. We select
\begin{equation}
\begin{gathered}
\mathcal{I}_{Q}^{(\ell)}
=\operatorname{TopK}(S_{\mathrm{head}}^{(\ell)},H_s),\\
S_{\mathrm{KV},k}^{(\ell)}
=\sum_{h\in\mathcal{G}(k)}S_{\mathrm{head},h}^{(\ell)},\\
\mathcal{I}_{\mathrm{KV}}^{(\ell)}
=\operatorname{TopK}(S_{\mathrm{KV}}^{(\ell)},H_s^{\mathrm{KV}}),
\end{gathered}
\label{eq:head-selection}
\end{equation}
where $\mathcal{G}(k)$ is the group of donor query heads sharing KV head $k$.
This mirrors the code's grouped-query structure rather than applying one head
index set to all attention projections.

The index collection
$\mathcal{S}_{\mathcal{D}}=
\{\mathcal{I}_{L},\mathcal{I}_{d},
\mathcal{I}_{Q}^{(\ell)},\mathcal{I}_{\mathrm{KV}}^{(\ell)},
\mathcal{I}_{m}^{(\ell)}\}_{\ell\in\mathcal{I}_{L}}$
defines the task-conditioned extraction operator:
\begin{equation}
\theta_l^{\mathrm{APM}}
=\mathcal{P}_{\mathcal{D}}
(\theta_l;\mathcal{S}_{\mathcal{D}},\mathcal{A}_s).
\label{eq:target-shape-extraction}
\end{equation}
Each index set is applied consistently to coupled tensors: hidden indices to
embeddings, normalization parameters, and projection input/output axes; query
heads to Q and output projections; KV heads to K/V projections; and MLP indices
to gate, up, and down projections. The result is a dense,
recipient-shaped donor slice that preserves tensor compatibility and is
constructed without optimization.

\subsection{Micro-Injection Merge}

Let $\theta_l^{\mathrm{APM}}$ denote the extracted donor slice.
Because it has the same tensor structure as $\theta_s$, APM uses a single linear micro-injection rule:
\begin{equation}
\theta_{\mathrm{APM}}
=(1-\mu)\theta_s+\mu\theta_l^{\mathrm{APM}},
\mu\in (0,0.10).
\label{eq:apm-merge}
\end{equation}
The recipient remains the dominant parameter source and retains its original inference architecture.
The small coefficient keeps the update in a local neighborhood of the recipient while allowing the selected donor slice to bias task-relevant computations.
Because the extraction step has already matched every tensor shape, the merge does not require adapters, projections, or further optimization.
Multiple extracted donors can be composed by assigning each a small coefficient; the corresponding formula is given in the supplementary material.
Algorithm~\ref{alg:apm} summarizes the complete framework.
\begin{algorithm}[t]
\caption{Activation-Prune-Merge}
\label{alg:apm}
\begin{algorithmic}[1]
\REQUIRE Donor $M_l$, recipient $M_s$, activation dataset $\mathcal{D}$,
profiling token rule $\Omega$, injection ratio $\mu$
\ENSURE Fused model $M_{\mathrm{APM}}$
\STATE Run $M_l$ on $\mathcal{D}$ and collect activation statistics over
$\Omega$
\STATE Compute layer, hidden-channel, attention-head, and MLP-neuron scores
\STATE Select depth by the specified strategy and activation-ranked hidden,
Q/KV-head, and MLP indices to match $\mathcal{A}_s$
\STATE $\theta_l^{\mathrm{APM}}\leftarrow
\mathcal{P}_{\mathcal{D}}(\theta_l;\mathcal{S}_{\mathcal{D}},\mathcal{A}_s)$
\STATE $\theta_{\mathrm{APM}}\leftarrow
(1-\mu)\theta_s+\mu\theta_l^{\mathrm{APM}}$
\STATE \textbf{return} $M_{\mathrm{APM}}=(\mathcal{A}_s,
\theta_{\mathrm{APM}})$
\end{algorithmic}
\end{algorithm}

\section{Experiments}

\begin{table}[!t]
\centering
\small
\renewcommand{\arraystretch}{1.22}
\begin{tabularx}{\columnwidth}{>{\raggedright\arraybackslash}X>{\raggedright\arraybackslash}X}
\hline
  \textbf{Family} & \textbf{Benchmarks} \\
\hline
Math & GSM8K, MATH-500 \\
Code generation & HumanEvalPlus, MBPPPlus \\
Instruction following & IFEval \\
QA and NLU & MNLI, RTE, QNLI, PIQA, Winogrande, COPA, BoolQ \\
Commonsense \& knowledge & ARC, HellaSwag, MMLU \\
Comprehensive reasoning & BBH \\
\hline
\end{tabularx}
\caption{Benchmark families used in our experiments.}
\label{tab:exp-datasets}
\end{table}

\begin{table}[!t]
\centering
\small
\setlength{\tabcolsep}{4pt}
\begin{tabularx}{\columnwidth}{@{}lcl>{\raggedright\arraybackslash}X@{}}
\toprule
\textbf{Role} & \textbf{Model} & \textbf{Scale} & \textbf{Use in APM} \\
\midrule
Recipient & Qwen2.5-3B  & 3B  & Target architecture and dominant merge component \\
Donor     & Qwen2.5-14B & 14B & Sequential multi-stage fusion analyses \\
Donor     & Qwen2.5-32B & 32B & Cross-scale activation-guided extraction \\
\bottomrule
\end{tabularx}
\caption{Model roles in our experiments.
Qwen2.5-32B is the donor used for the primary benchmark results; Qwen2.5-14B is used in the sequential multi-stage fusion analysis.}
\label{tab:model-roles}
\end{table}

\begin{table}[!t]
\centering
\small
\setlength{\tabcolsep}{0pt}
\renewcommand{\arraystretch}{1.08}
\begin{tabular*}{\columnwidth}{@{\extracolsep{\fill}}lrrrrc@{}}
\toprule
 Dataset & 3B (\%) & IM (\%) & APM (\%) & $\Delta$3B (pp) & $\Delta$IM (pp) \\
\midrule
GSM8K & 65.7 & 72.2 & \textbf{72.4} & +6.7 & +0.2 \\
MATH-500 & 38.4 & 40.6 & \textbf{42.8} & +4.4 & +2.2 \\
HumanEvalPlus & 39.6 & 38.4 & \textbf{40.9} & +1.3 & +2.5 \\
MBPPPlus & 57.9 & 59.8 & \textbf{61.4} & +3.5 & +1.6 \\
IFEval & 40.7 & 40.8 & \textbf{43.1} & +2.4 & +2.3 \\
MNLI & 46.0 & 53.3 & \textbf{57.0} & +11.0 & +3.7 \\
RTE & 64.3 & 74.4 & \textbf{82.3} & +18.0 & +7.9 \\
QNLI & 52.3 & 55.3 & \textbf{65.7} & +13.4 & +10.4 \\
PIQA & 76.1 & 76.1 & \textbf{76.7} & +0.6 & +0.6 \\
Winogrande & 62.8 & 63.5 & \textbf{64.6} & +1.8 & +1.1 \\
COPA & 77.0 & 80.0 & \textbf{81.0} & +4.0 & +1.0 \\
BoolQ & 70.8 & 72.7 & \textbf{79.2} & +8.4 & +6.5 \\
ARC & 39.4 & \textbf{39.9} & \textbf{39.9} & +0.5 & 0.0 \\
HellaSwag & \textbf{67.9} & 67.1 & 67.6 & -0.3 & +0.5 \\
MMLU & 45.0 & 45.0 & \textbf{45.6} & +0.6 & +0.6 \\
BBH & 43.6 & 45.6 & \textbf{49.9} & +6.3 & +4.3 \\
\midrule
Avg. & 55.5 & 57.8 & \textbf{60.6} & +5.1 & +2.8 \\
\bottomrule
\end{tabular*}
\caption{Main results on 16 benchmarks using Qwen2.5-3B as the recipient and Qwen2.5-32B as the donor.
APM denotes Activation-Prune-Merge, and IM denotes Intersection-Merge, the baseline introduced by \citet{fan2026rethinkingheterogeneousllmmerging}.
Scores are percentages, and differences are reported in percentage points (pp).
Higher is better for all metrics.}
\label{tab:main-results}
\end{table}

\subsection{Benchmark Datasets}

We evaluate APM on 16 benchmarks across six capability families (Table~\ref{tab:exp-datasets}).
GSM8K and the MATH dataset cover mathematical reasoning, while the MATH-500 evaluation subset follows the representative MATH-test subset introduced for process-supervised verification \citep{cobbe2021gsm8k,hendrycks2021math,lightman2023verify}; HumanEvalPlus and MBPPPlus assess code generation \citep{liu2023evalplus}; and IFEval measures instruction following \citep{zhou2023ifeval}.
The QA and NLU suite contains MNLI, RTE, QNLI, PIQA, WinoGrande, COPA, and BoolQ \citep{williams2018mnli,dagan2006rte,wang2018glue,bisk2020piqa, sakaguchi2020winogrande,roemmele2011copa,clark2019boolq}.
We further use ARC, HellaSwag, and MMLU for commonsense and broad-domain knowledge, and BBH for challenging multi-task reasoning \citep{clark2018arc,zellers2019hellaswag,hendrycks2021mmlu,suzgun2022bbh}.

\subsection{Models}

We use Qwen2.5-3B as the recipient and Qwen2.5-14B and Qwen2.5-32B as cross-scale donors, all from the Qwen2.5 model family described in the technical report \citep{qwen25technical}.
Table~\ref{tab:model-roles} summarizes these roles.
We consistently use the terms \emph{recipient} and \emph{donor} to denote model roles.
APM performs direct, training-free parameter transfer.
Each donor is activation-profiled, pruned to the Qwen2.5-3B architecture, and then micro-injected, allowing us to test transfer and composition across donor scales.

\subsection{Experimental Setup}

As summarized in Figure~\ref{fig:activation-pruning-overview}, we construct a task-specific activation set from each benchmark and profile the corresponding donor before extraction.
For generative tasks, we consider three profiling scopes: \emph{prompt-only}, \emph{generation-only}, and \emph{prompt--generation}.
Generated continuations are capped at a task-appropriate length.
For multiple-choice tasks, we analogously use \emph{prompt-only}, \emph{choice-only}, or \emph{prompt--choice} profiling.
These alternatives control which token positions contribute to the activation statistics while preserving the original model parameters during profiling.
The activation examples used for donor profiling are kept disjoint from the held-out examples used for evaluation.

After activation profiling, we compare three target-depth layer-selection rules.
\emph{First-$n$} retains the first $n$ donor layers, \emph{uniform} selects $n$ approximately evenly spaced layers over the donor depth, and \emph{activation-guided} retains the $n$ layers with the highest activation scores while preserving their original order.
In every case, $n$ is fixed by the recipient depth.
The selected donor is subsequently reduced to the full recipient shape and merged using the same micro-injection rule.

\subsection{Baselines}

We compare APM with the original recipient and IM \citep{fan2026rethinkingheterogeneousllmmerging}.
IM performs deterministic front-aligned extraction: it retains the first $L_s$ donor layers and, for each weight tensor, the leading indices along every dimension required by the recipient shape.
The resulting recipient-shaped donor slice is injected into the recipient with the same mixing coefficient used by APM.
This controlled comparison isolates the effect of activation-guided donor selection from the linear merge rule itself.

\paragraph{Overall results.} Table~\ref{tab:main-results} provides a complete overview across all 16 benchmarks.
APM increases the average score of the original Qwen2.5-3B recipient from 55.5\% to 60.6\%.
Under the same cross-scale fusion setting, APM exceeds or matches IM on every benchmark and raises its average from 57.8\% to 60.6\%.
The improvements span mathematics, code generation, instruction following, natural language understanding, and comprehensive reasoning, with particularly large gains on RTE ($+18.0$ pp over 3B and $+7.9$ pp over IM), QNLI ($+13.4$ and $+10.4$ pp), and BoolQ ($+8.4$ and $+6.5$ pp).
Additional significance analysis in the supplementary material is consistent with these gains.
On the eight selected BBH subtasks, the task-matched APM slices improve over IM on every task (Table~\ref{tab:bbh-subtasks}(a)). The largest gains occur for T2 ($+30.0$ pp), T1 ($+17.6$ pp), and T8 ($+12.8$ pp). The exact permutation tests in Table~\ref{tab:bbh-subtasks}(b) further support a positive matched-task advantage in both raw and standardized analyses.
These broad and consistent gains indicate that activation-guided extraction produces a more effective donor slice than front-aligned intersection under the same micro-injection framework.

\section{Analysis}

\begin{table}[t]
\centering
\small
\setlength{\tabcolsep}{1pt}
\renewcommand{\arraystretch}{1.08}
\textbf{(a) Cross-task accuracy matrix}\\[2pt]
\begin{tabular*}{\columnwidth}{@{\extracolsep{\fill}}lrrrrrrrrr@{}}
\toprule
Model & T1 & T2 & T3 & T4 & T5 & T6 & T7 & T8 & Avg. (\%) \\
\midrule
Qwen2.5-3B  & 32.4 & 14.4 & 37.0 &  6.2 & 32.0 & 28.8 & 22.0 & 45.2 & 27.3 \\
Qwen2.5-32B & 79.2 & 74.0 & 91.8 & 78.7 & 98.0 & 99.6 & 97.6 & 98.0 & 89.6 \\
IM          & 35.6 & 25.6 & 57.5 &  7.3 & 31.2 & 30.8 & 24.4 & 44.4 & 32.1 \\
\midrule
APM-T1 & \underline{\textbf{53.2}} & 54.4 & 61.0 &  9.0 & 30.8 & 35.2 & 28.4 & 58.8 & 41.3 \\
APM-T2 & 48.0 & \underline{\textbf{55.6}} & 61.6 & 11.2 & \textbf{32.4} & 36.8 & \textbf{31.2} & \textbf{60.0} & \textbf{42.1} \\
APM-T3 & 45.6 & 46.0 & \underline{61.6} & 10.7 & \textbf{32.4} & 37.6 & 27.6 & 57.6 & 39.9 \\
APM-T4 & 49.2 & 49.6 & \textbf{63.0} & \underline{\textbf{15.7}} & 30.8 & 35.2 & 30.0 & 55.2 & 41.1 \\
APM-T5 & 45.2 & 50.0 & 61.6 & 12.4 & \underline{32.0} & 35.6 & 28.0 & 57.6 & 40.3 \\
APM-T6 & 47.6 & 51.2 & 60.3 & 11.8 & 31.2 & \underline{38.0} & 30.0 & 57.6 & 41.0 \\
APM-T7 & 43.6 & 49.6 & 60.3 &  9.0 & 32.0 & \textbf{39.6} & \underline{28.4} & 56.4 & 39.9 \\
APM-T8 & 50.0 & 49.6 & \textbf{63.0} & 13.5 & 31.6 & 38.4 & 30.0 & \underline{57.2} & 41.7 \\
\bottomrule
\end{tabular*}
\\[6pt]
\renewcommand{\arraystretch}{1.10}
\setlength{\tabcolsep}{4pt}
\textbf{(b) Exact permutation tests}\\[2pt]
\begin{tabular*}{\columnwidth}{@{\extracolsep{\fill}}llrrrr@{}}
\toprule
Suite & Scale & Diag. & Off-diag. & $\Delta$ & $p$ \\
\midrule
BBH-8 & Raw & 42.7 & 40.6 & 2.1 & 0.0028 \\
BBH-8 & Standardized & 0.734 & -0.105 & 0.839 & 0.0198 \\
\bottomrule
\end{tabular*}
\\[6pt]
\textbf{(c) Transfer to the complete BBH-27 suite}\\[2pt]
\setlength{\tabcolsep}{2pt}
\begin{tabular*}{\columnwidth}{@{\extracolsep{\fill}}lrrrrr@{}}
\toprule
Model & Avg. (\%) & $\Delta$3B & $\Delta$IM & Wins/3B & Wins/IM \\
\midrule
Qwen2.5-3B & 43.6 & -- & -- & -- & -- \\
IM & 45.3 & +1.6 & -- & 15/27 & -- \\
\midrule
APM-T1 & 48.3 & +4.7 & +3.0 & 18/27 & 18/27 \\
APM-T2 & 49.7 & +6.0 & +4.4 & 19/27 & 20/27 \\
APM-T3 & 48.9 & +5.2 & +3.6 & 19/27 & \textbf{21/27} \\
APM-T4 & \textbf{49.9} & \textbf{+6.3} & \textbf{+4.6} & 17/27 & 19/27 \\
APM-T5 & 49.8 & +6.2 & +4.5 & 19/27 & 20/27 \\
APM-T6 & 49.4 & +5.7 & +4.1 & 19/27 & \textbf{21/27} \\
APM-T7 & 48.4 & +4.8 & +3.2 & \textbf{21/27} & 20/27 \\
APM-T8 & 49.6 & +6.0 & +4.3 & 19/27 & 20/27 \\
\bottomrule
\end{tabular*}
\caption{Fine-grained BBH transfer and permutation analysis. Panel~(a) reports BBH-8 cross-task accuracies; Panel~(b) reports exact tests of diagonal advantage; Panel~(c) evaluates transfer of slices T1--T8 on the complete BBH-27 suite.}
\label{tab:bbh-subtasks}
\end{table}

\begin{table}[!t]
\centering
\small
{
\setlength{\tabcolsep}{1.8pt}
\renewcommand{\arraystretch}{1.04}
\setlength{\aboverulesep}{1pt}
\setlength{\belowrulesep}{1pt}
\textbf{(a) IM and APM across injection ratios}\\[2pt]
\begin{tabular*}{\columnwidth}{@{\extracolsep{\fill}}crrrrrrrr@{}}
\toprule
& \multicolumn{2}{c}{BoolQ} & \multicolumn{2}{c}{MNLI} & \multicolumn{2}{c}{QNLI} & \multicolumn{2}{c}{RTE} \\[-2pt]
\cmidrule(lr){2-3}\cmidrule(lr){4-5}\cmidrule(lr){6-7}\cmidrule(lr){8-9}
$\mu$ & IM & APM & IM & APM & IM & APM & IM & APM \\[-1pt]
\midrule
0.01 & 70.9 & 71.4 & 48.2 & 50.5 & 53.5 & 53.2 & 69.3 & 70.4 \\
0.02 & 70.4 & 75.3 & 49.2 & 52.2 & 52.8 & 55.0 & 72.6 & 75.8 \\
0.03 & 71.2 & 76.7 & 50.4 & 55.4 & 53.8 & 57.4 & 75.8 & 79.8 \\
0.04 & 72.5 & 79.0 & 52.0 & 56.8 & 54.5 & 63.5 & 77.3 & 81.6 \\
0.05 & 72.7 & 79.2 & 53.3 & \textbf{57.0} & 55.3 & 65.7 & 74.4 & \textbf{82.3} \\
0.06 & 73.6 & \textbf{79.6} & 52.9 & 56.6 & 56.3 & 73.1 & 73.7 & 80.9 \\
0.08 & 75.0 & 77.9 & 49.0 & 52.2 & 60.9 & \textbf{73.9} & 65.3 & 73.3 \\
0.10 & 75.8 & 75.2 & 46.2 & 45.7 & 55.9 & 52.1 & 60.3 & 59.2 \\
\bottomrule
\end{tabular*}
}
\\[3pt]
\small
\renewcommand{\arraystretch}{1.04}
\textbf{(b) Recipient and donor reference scores}\\[2pt]
\setlength{\tabcolsep}{7pt}
\begin{tabular*}{\columnwidth}{@{\extracolsep{\fill}}lrrrr@{}}
\toprule
Model & BoolQ & MNLI & QNLI & RTE \\
\midrule
Qwen2.5-3B  & 70.7 & 46.1 & 52.9 & 63.5 \\
Qwen2.5-32B & 89.0 & 66.9 & 84.2 & 80.9 \\
\bottomrule
\end{tabular*}
\caption{Accuracy across the low-to-moderate injection region.
Panel (a) reports IM and task-activated APM under the same injection ratio; bold entries are the best APM scores within the displayed interval.
Panel (b) gives the Qwen2.5-3B recipient and 32B donor references.
Scores are percentages.}
\label{tab:ratio-low-region}
\end{table}

\begin{figure}[!t]
\centering
\includegraphics[width=\columnwidth]{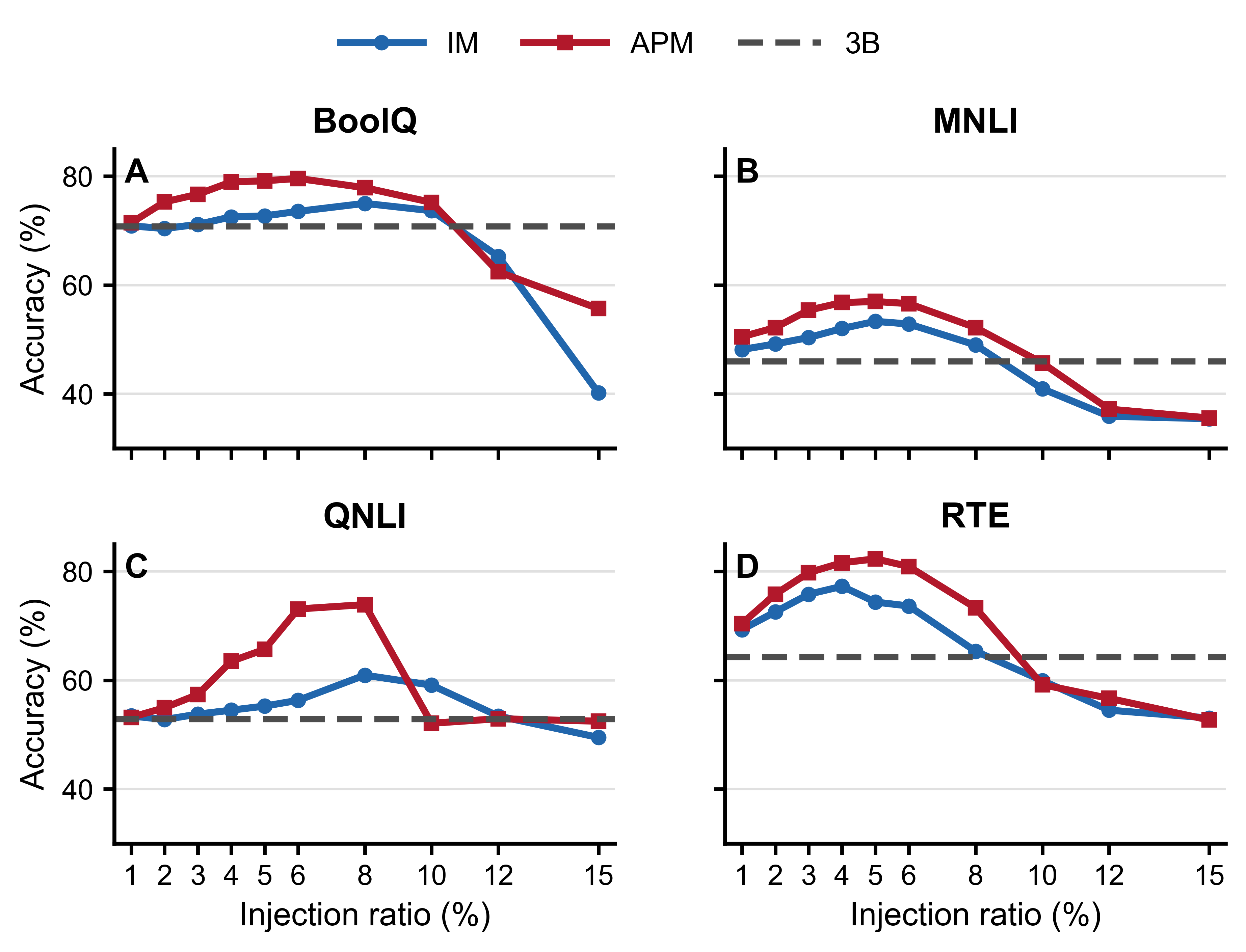}
\caption{Accuracy as a function of the injection ratio on four benchmarks.}
\label{fig:ratio-sensitivity}
\end{figure}

\begin{table}[t]
\centering
\small
\setlength{\tabcolsep}{3pt}
\renewcommand{\arraystretch}{1.04}
\textbf{(a) Task-wise accuracy across fusion stages}\\[2pt]
\begin{tabular*}{\columnwidth}{@{\extracolsep{\fill}}lrrrrr@{}}
\toprule
Task & 3B & IM-S1 & IM-S2 & APM-S1 & APM-S2 \\
\midrule
MNLI & 46.1 & 49.3 & 52.7 & 52.2 & \textbf{55.0} \\
RTE & 63.5 & 72.2 & 75.5 & 75.5 & \textbf{80.1} \\
QNLI & 52.8 & 52.7 & 52.3 & 55.4 & \textbf{59.3} \\
WinoGrande & 63.1 & \textbf{63.5} & 62.9 & 62.7 & 62.7 \\
PIQA & 75.7 & 76.1 & 75.4 & \textbf{76.7} & 75.5 \\
BoolQ & 70.7 & 70.9 & 69.6 & 74.0 & \textbf{75.7} \\
COPA & 78.0 & 80.0 & 79.0 & \textbf{81.0} & \textbf{81.0} \\
\midrule
Avg. & 64.3 & 66.4 & 66.7 & 68.2 & \textbf{69.9} \\
\bottomrule
\end{tabular*}
\\[3pt]
\textbf{(b) Average gain at each fusion hop}\\[2pt]
\setlength{\tabcolsep}{4pt}
\begin{tabular*}{\columnwidth}{@{\extracolsep{\fill}}lrrrrr@{}}
\toprule
Method & Base & S1 & $\Delta_1$ (pp) & S2 & $\Delta_2$ (pp) \\
\midrule
IM & 64.3 & 66.4 & +2.1 & 66.7 & +0.3 \\
APM & 64.3 & 68.2 & \textbf{+3.9} & \textbf{69.9} & \textbf{+1.7} \\
\bottomrule
\end{tabular*}
\caption{Sequential two-stage fusion across seven QA and NLU benchmarks.
Panel~(a) reports task-wise accuracy, and Panel~(b) summarizes the average gain at each fusion hop.
Stage~1 (S1) injects the 32B donor slice into the 3B recipient, and Stage~2 (S2) injects the 14B slice into the S1 model.
Each stage uses $\mu=0.02$.
Accuracies are percentages, and gains are percentage points (pp).}
\label{tab:multi-source}
\end{table}

\begin{table}[!t]
\centering
\small
\renewcommand{\arraystretch}{1.04}
\begin{tabular*}{\columnwidth}{@{\extracolsep{\fill}}lcc@{}}
\toprule
\multicolumn{3}{@{}l}{(a) Activation-data selection} \\
\addlinespace[2pt]
Activation-data selection & Mean (\%) & Norm. var. \\
\midrule
Random & 56.9 & 0.0090 \\
32B correct / 3B wrong & \textbf{58.5} & 0.0112 \\
32B correct / 3B correct & \textbf{58.5} & 0.0100 \\
\midrule
\multicolumn{3}{@{}l}{(b) Activation-set size} \\
\addlinespace[2pt]
Examples & Mean (\%) & Norm. var. \\
\midrule
25 & \textbf{59.6} & 0.0106 \\
50 & 57.4 & 0.0120 \\
100 & 58.3 & 0.0089 \\
200 & 58.6 & 0.0114 \\
500 & 56.5 & 0.0076 \\
1000 & 57.3 & 0.0091 \\
\bottomrule
\end{tabular*}
\caption{Mean QNLI accuracy and normalized variance grouped by activation-data selection strategy and activation-set size.
Accuracy scores are percentages.}
\label{tab:activation-data}
\end{table}

\subsection{Task-Matched Transfer}

We examine eight BBH subtasks with substantial transfer headroom: the Qwen2.5-3B student averages 27.3\%, whereas the Qwen2.5-32B teacher reaches 89.6\% (Table~\ref{tab:bbh-subtasks}(a)).
At the matched injection ratio $\mu=0.02$, IM raises the average to 32.1\%, providing a controlled reference that isolates donor-slice selection from the injected parameter mass.
All eight APM slices average 39.9--42.1\%, exceeding both baselines.
Matched activation improves over IM on every subtask, by 0.8--30.0 pp.
Several off-diagonal entries are also column maxima, indicating that activation-guided selection captures donor components with cross-task value rather than only task-specific structure.

This transfer persists across the complete BBH-27 suite (Table~\ref{tab:bbh-subtasks}(c)).
Panel~(c) reports average accuracies in percentages, gains over the 3B recipient and IM in percentage points, and the number of BBH tasks on which each APM model exceeds the corresponding baseline; T1--T8 follow the activation-source mapping in Panel~(a).
The supplementary material reports the corresponding per-task accuracies for all 27 BBH subtasks, together with the number of evaluation examples in each subtask.
The eight APM models average 48.3--49.9\%, gaining 4.7--6.3 pp over the student and 3.0--4.6 pp over IM.
Each APM variant also outperforms the student on 17--21 of the 27 tasks and IM on 18--21 tasks. Thus, the gains are not confined to the subtask used as the activation source: a donor slice extracted from one activation group can improve accuracy across many other BBH tasks. This broad transfer indicates that activation-guided pruning captures donor components with general cross-task value, rather than only information specialized to the activation-source task.

Panel~(b) reports raw accuracies and column-normalized APM gains over IM; Diag., Off-diag., $\Delta$, and $p$ denote the matched mean, unmatched mean, their difference, and the one-sided positive-advantage probability.
Both the raw-score and normalized-gain tests show a significant matched-task advantage, indicating that matched activation profiling usually yields the strongest donor slice, while several off-diagonal maxima still show effective cross-task transfer (Table~\ref{tab:bbh-subtasks}(b)).

\subsection{Injection-Ratio Sensitivity}

Following the IM evaluation over ratios from $0.01$ to $0.05$ \citep{fan2026rethinkingheterogeneousllmmerging}, we examine a finer range on BoolQ, MNLI, QNLI, and RTE to test robustness to the injected donor mass (Table~\ref{tab:ratio-low-region} and Figure~\ref{fig:ratio-sensitivity}).
APM and IM show similar ratio-dependent trends, but APM generally performs better throughout the effective region $(0,0.10)$ under the same interpolation rule.
Notably, APM reaches 82.3\% on RTE at $\mu=0.05$, surpassing the 32B donor's 80.9\%, and peaks at 73.9\% on QNLI at $\mu=0.08$.
The optimal ratio varies by task, suggesting that task-aware ratio selection can further improve activation-guided extraction.
Results through $\mu=0.15$ are reported in Supplementary Table~1.

\subsection{Sequential Multi-Stage Fusion}
We test sequential fusion by injecting 32B and 14B donor slices into the 3B recipient in two stages, each at $\mu=0.02$, using the same order and coefficients for IM (Table~\ref{tab:multi-source}).
APM raises the seven-task average from 64.3\% to 68.2\% and then 69.9\%, whereas IM reaches 66.4\% and 66.7\%.
The second-stage gain is therefore 1.7 pp for APM versus 0.3 pp for IM.
APM-S2 achieves the highest scores on MNLI, RTE, QNLI, BoolQ, and COPA, while APM-S1 leads on PIQA.
These results indicate that activation-guided donor slices remain compatible with repeated small-ratio fusion.

\subsection{Sensitivity to Activation-Data Selection}

On QNLI at $\mu=0.05$, donor-correct subsets average 58.5\%, compared with 56.9\% for random selection across nine activation--pruning variants (Table~\ref{tab:activation-data}).
Thus, donor correctness is more informative than recipient difficulty alone, and recipient-solved examples remain useful when the donor is also correct.
The 25-example set performs best at 59.6\%, versus 58.6\% for 200 examples and 57.3\% for 1,000, so more activation data is not necessarily better.
Complete results are provided in the supplementary material.

\section{Conclusion}

We introduced Activation-Prune-Merge (APM), a training-free extension of cross-scale micro-injection that replaces fixed donor truncation with task-conditioned activation profiling and structured component selection.
Across diverse benchmark families, APM consistently strengthens the original recipient and improves over IM under the same interpolation rule.
Its fine-grained BBH transfer, broad effectiveness across small injection ratios, and continued gains under sequential fusion further demonstrate the quality and composability of the extracted donor slices.

The strength of APM lies in its simplicity.
By adding lightweight activation statistics and target-shape component selection to the IM pipeline, APM obtains substantial improvements while preserving the recipient architecture, inference cost, and training-free workflow.
The large gains produced by this lightweight extension demonstrate the effectiveness of activation-guided donor slice selection for cross-scale transfer.
More broadly, our results reveal substantial untapped potential for directly fusing heterogeneous models without explicit neuron-wise semantic alignment, opening a practical path for strong models to transfer useful capabilities into compact recipients.

Taken together, these results position APM as a lightweight complement to model scaling.
Its separation of target-shape reduction and activation-guided selection keeps the recipient architecture and micro-injection rule unchanged, facilitating adaptation across benchmark families.



\long\def\SupplementaryText{%
\section{Detailed APM Formulation}
\label{app:apm-details}

For each input $x\in\mathcal{D}$, let $\Omega(x)$ denote the token positions used for profiling and let $Z=\sum_{x\in\mathcal{D}}|\Omega(x)|$.
For donor layer $r$, hidden channel $i$, attention head $k$, and MLP neuron $j$, APM uses the following activation scores:
\begin{align}
s_r^{\mathrm{layer}}
&=\frac{1}{Z}\sum_{x\in\mathcal{D}}
  \sum_{t\in\Omega(x)}\lVert h_r(x,t)\rVert_2, \\
s_i^{\mathrm{hidden}}
&=\frac{1}{Z}\sum_{x\in\mathcal{D}}
  \sum_{t\in\Omega(x)}|h_i(x,t)|, \\
s_{r,k}^{\mathrm{head}}
&=\frac{1}{Z}\sum_{x\in\mathcal{D}}
  \sum_{t\in\Omega(x)}\lVert o_{r,k}(x,t)\rVert_2, \\
s_{r,j}^{\mathrm{mlp}}
&=\frac{1}{Z}\sum_{x\in\mathcal{D}}
  \sum_{t\in\Omega(x)}|a_{r,j}(x,t)|.
\label{eq:activation-scores}
\end{align}
Here, $h_r$ is the layer output, $h_i$ is a hidden-channel activation, $o_{r,k}$ is the output of attention head $k$, and $a_{r,j}$ is the post-activation value of MLP neuron $j$.
If the recipient requires $L_s$ layers, hidden width $d_s$, attention
configuration $H_s$, and MLP width $m_s$, the selected index sets are
\begin{equation}
\begin{aligned}
\mathcal{I}_{L}^{*}&=\operatorname{TopK}(S_{\mathrm{layer}},L_s), &
\mathcal{I}_{d}^{*}&=\operatorname{TopK}(S_{\mathrm{hidden}},d_s),\\
\mathcal{I}_{H}^{*}&=\operatorname{TopK}(S_{\mathrm{head}},H_s), &
\mathcal{I}_{m}^{*}&=\operatorname{TopK}(S_{\mathrm{mlp}},m_s).
\end{aligned}
\label{eq:topk-selection}
\end{equation}
Writing their collection as $\mathcal{S}^{*}$, structured extraction is
\begin{equation}
\theta_l^{\mathrm{APM}}
=\mathcal{P}(\theta_l;\mathcal{S}^{*},\mathcal{A}_s).
\label{eq:apm-pruning}
\end{equation}

For $K$ donor slices with coefficients $\{\mu_q\}_{q=1}^{K}$ satisfying
$\sum_q\mu_q<1$, multi-donor micro-injection is
\begin{equation}
\theta_{\mathrm{multi}}
=\left(1-\sum_{q=1}^{K}\mu_q\right)\theta_s
 +\sum_{q=1}^{K}\mu_q\theta_{l_q}^{\mathrm{APM}}.
\label{eq:multi-donor}
\end{equation}

\section{Statistical Analysis of Diagonal Effects}
\label{app:diagonal-test}

Let $S=(S_{ij})_{i,j=1}^{K}$ be a cross-task accuracy matrix whose rows index activation sources and whose columns index evaluation tasks in the same order.
We quantify matched-task performance by the diagonal mean
\begin{equation}
D(S)=\frac{1}{K}\sum_{i=1}^{K}S_{ii},
\end{equation}
and unmatched-task performance by the mean of all off-diagonal entries
\begin{equation}
O(S)=\frac{1}{K(K-1)}\sum_{i=1}^{K}\sum_{j\ne i}S_{ij}.
\end{equation}
The observed diagonal advantage is
\begin{equation}
T_{\mathrm{obs}}=D(S)-O(S).
\label{eq:diagonal-statistic}
\end{equation}

\paragraph{Exact permutation test.} Under the null hypothesis that activation-source labels are exchangeable with respect to evaluation-task labels, we enumerate every permutation $\pi\in\mathcal{S}_K$.
For each permutation, the entries treated as matched are $\{S_{\pi(j),j}\}_{j=1}^{K}$.
We compute
\begin{align}
D_{\pi}(S)
&=\frac{1}{K}\sum_{j=1}^{K}S_{\pi(j),j},\\
O_{\pi}(S)
&=\frac{\sum_{i,j}S_{ij}-\sum_j S_{\pi(j),j}}{K(K-1)},\\
T_{\pi}
&=D_{\pi}(S)-O_{\pi}(S).
\end{align}
The one-sided exact $p$-value for a positive diagonal advantage is
\begin{equation}
p_{+}=\frac{1}{K!}\sum_{\pi\in\mathcal{S}_K}
\mathbf{1}\!\left[T_{\pi}\ge T_{\mathrm{obs}}\right].
\label{eq:permutation-p}
\end{equation}
Because the permutation spaces are small, we use exhaustive enumeration rather than Monte Carlo sampling: $8!=40{,}320$ permutations for BBH-8 and $7!=5{,}040$ for QA-7.

\paragraph{IM-relative gains and column standardization.} Let $b_j$ be the IM score on evaluation task $j$.
We first express each APM result as its gain over IM,
\begin{equation}
\Delta_{ij}=S_{ij}-b_j.
\end{equation}
Subtracting a common baseline within each column improves interpretability but does not change the unstandardized test statistic: both the diagonal and off-diagonal means decrease by $K^{-1}\sum_j b_j$, so $T(\Delta)=T(S)$.
To give equal scale to tasks with different across-source variation, we additionally compute column-wise z-scores
\begin{equation}
Z_{ij}=\frac{\Delta_{ij}-\bar{\Delta}_{\cdot j}}{\sigma_j},
\qquad
\bar{\Delta}_{\cdot j}=\frac{1}{K}\sum_{i=1}^{K}\Delta_{ij},
\end{equation}
where
\begin{equation}
\sigma_j=
\sqrt{\frac{1}{K}\sum_{i=1}^{K}
\left(\Delta_{ij}-\bar{\Delta}_{\cdot j}\right)^2}.
\end{equation}
We then apply Equations~\ref{eq:diagonal-statistic}--\ref{eq:permutation-p} to $Z$.
This standardized test asks whether matched activation is advantageous relative to the typical variation among activation sources for each evaluation task, rather than being driven by tasks with larger absolute score ranges.

}

\long\def\MathSignificanceTable{%
\begingroup
\section{MATH-500 Significance Test}
\vspace{10pt}
\begin{center}
\centering
\small
\setlength{\tabcolsep}{4pt}
\renewcommand{\arraystretch}{1.12}
\begin{minipage}[t]{0.38\textwidth}
\centering
\begin{tabular}{lcc}
\toprule
\multicolumn{3}{c}{\textbf{(a) Accuracy}} \\
\midrule
Model & Accuracy & Lower CI \\
\midrule
3B & 0.3840 & 0.3420 \\
IM & 0.3940 & 0.3520 \\
APM & \textbf{0.4280} & 0.3840 \\
\bottomrule
\end{tabular}
\end{minipage}
\hfill
\begin{minipage}[t]{0.54\textwidth}
\centering
\begin{tabular}{lccc}
\toprule
\multicolumn{4}{c}{\textbf{(b) Pairwise differences}} \\
\midrule
Comparison & Difference & Lower CI & $p$ \\
\midrule
IM $-$ 3B & +0.0100 & -0.0220 & 0.5723 \\
APM $-$ 3B & \textbf{+0.0440} & \textbf{0.0120} & \textbf{0.0073} \\
APM $-$ IM & \textbf{+0.0340} & \textbf{0.0060} & \textbf{0.0195} \\
\bottomrule
\end{tabular}
\end{minipage}
\captionof{table}{MATH-500 significance test under \texttt{math\_verify}.~The upper panel reports accuracy and the lower confidence-interval bound. The lower panel reports pairwise accuracy differences, lower confidence-interval bounds, and $p$-values. Bold entries indicate significant APM improvements at $p<0.05$.}
\label{tab:math500-significance}
\end{center}
\endgroup
}

\long\def\RatioTable{%
\begingroup
\section{Complete Injection-Ratio Results}
\vspace{4pt}
\centering
\small
\setlength{\tabcolsep}{3.2pt}
\renewcommand{\arraystretch}{1.03}
\begin{tabular*}{\textwidth}{@{\extracolsep{\fill}}c*{8}{r}@{}}
\toprule
& \multicolumn{2}{c}{BoolQ} & \multicolumn{2}{c}{MNLI} & \multicolumn{2}{c}{QNLI} & \multicolumn{2}{c}{RTE} \\
\cmidrule(lr){2-3}\cmidrule(lr){4-5}\cmidrule(lr){6-7}\cmidrule(l){8-9}
Injection ratio $\mu$ & IM & APM & IM & APM & IM & APM & IM & APM \\
\midrule
0.01 & 70.89 & 71.44 & 48.17 & 50.51 & 53.45 & 53.19 & 69.31 & 70.40 \\
0.02 & 70.40 & 75.29 & 49.19 & 52.20 & 52.79 & 54.95 & 72.56 & 75.81 \\
0.03 & 71.16 & 76.70 & 50.37 & 55.40 & 53.78 & 57.39 & 75.81 & 79.78 \\
0.04 & 72.54 & 78.96 & 52.03 & 56.80 & 54.51 & 63.50 & 77.26 & 81.59 \\
0.05 & 72.72 & 79.17 & 53.34 & \textbf{56.99} & 55.26 & 65.68 & 74.37 & \textbf{82.31} \\
0.06 & 73.55 & \textbf{79.60} & 52.86 & 56.60 & 56.31 & 73.10 & 73.65 & 80.90 \\
0.08 & 75.02 & 77.90 & 49.01 & 52.20 & 60.94 & \textbf{73.90} & 65.34 & 73.30 \\
0.10 & 73.67 & 75.20 & 40.96 & 45.70 & 59.11 & 52.10 & 59.93 & 59.20 \\
\midrule
0.12 & 65.29 & 62.45 & 35.91 & 37.22 & 53.43 & 52.94 & 54.51 & 56.68 \\
0.15 & 40.21 & 55.69 & 35.46 & 35.59 & 49.50 & 52.48 & 53.07 & 52.71 \\
\bottomrule
\end{tabular*}
\setcounter{table}{0}
\captionof{table}{Complete injection-ratio results on four benchmarks. IM and task-activated APM are evaluated at the same ratios. Rows above the internal rule correspond to the effective interval reported in the main paper; the final two rows show the higher-ratio region in which the benefits weaken or disappear. Bold entries are the best APM scores for each benchmark across all evaluated ratios. Scores are percentages.}
\label{tab:complete-ratio}
\endgroup
}

\long\def\ActivationDataTable{%
\begingroup
\section{Detailed Activation-Data Selection Results}
\vspace{4pt}
\centering
\scriptsize
\setlength{\tabcolsep}{3.5pt}
\renewcommand{\arraystretch}{1.08}
\textbf{(a) Maximum QNLI accuracy across nine variants per condition}\par\vspace{3pt}
\begin{tabular*}{\textwidth}{@{\extracolsep{\fill}}lrrrrrr@{}}
\toprule
Activation-data selection & 25 & 50 & 100 & 200 & 500 & 1000 \\
\midrule
Random                     & 73.77 & 65.60 & 69.43 & 61.30 & 62.58 & 66.41 \\
32B correct / 3B wrong     & 68.95 & 74.89 & 64.67 & 73.51 & 65.17 & 67.27 \\
32B correct / 3B correct   & 67.95 & 73.99 & 70.82 & 71.90 & 66.23 & 66.94 \\
\bottomrule
\end{tabular*}

\vspace{8pt}
\textbf{(b) Mean QNLI accuracy / normalized variance across nine variants per condition}\par\vspace{3pt}
\setlength{\tabcolsep}{2.0pt}
\begin{tabular*}{\textwidth}{@{\extracolsep{\fill}}lcccccc@{}}
\toprule
Activation-data selection & 25 & 50 & 100 & 200 & 500 & 1000 \\
\midrule
Random
  & 58.43 / 0.0132 & 56.76 / 0.0100 & 58.43 / 0.0093 & 55.21 / 0.0038 & 54.81 / 0.0057 & 57.88 / 0.0072 \\
32B correct / 3B wrong
  & 61.42 / 0.0091 & 57.63 / 0.0129 & 57.79 / 0.0060 & 60.69 / 0.0144 & 56.26 / 0.0087 & 57.32 / 0.0096 \\
32B correct / 3B correct
  & 59.05 / 0.0081 & 57.92 / 0.0126 & 58.66 / 0.0111 & 59.98 / 0.0098 & 58.53 / 0.0064 & 56.62 / 0.0103 \\
\bottomrule
\end{tabular*}
\setcounter{table}{1}
\captionof{table}{Complete QNLI results by activation-data selection strategy and activation-set size. Panel~(a) reports the maximum accuracy among the nine activation--pruning variants in each condition. Panel~(b) reports the corresponding mean accuracy and normalized variance. Accuracy scores are percentages; activation-set sizes are numbers of examples. All results use an injection ratio of $\mu=0.05$.}
\label{tab:activation-data-complete}
\endgroup
}

\long\def\BBHTable{%
\begingroup
\section{Complete BBH-27 Transfer Results}
\vspace{4pt}
\centering
\scriptsize
\setlength{\tabcolsep}{1.2pt}
\renewcommand{\arraystretch}{0.96}
\begin{tabular*}{\textwidth}{@{\extracolsep{\fill}}>{\raggedright\arraybackslash}p{0.205\textwidth}r*{10}{r}@{}}
\toprule
BBH task & $N$ & 3B & IM & T1 & T2 & T3 & T4 & T5 & T6 & T7 & T8 \\
\midrule
Boolean expressions & 250 & 85.60 & 85.20 & 83.60 & 84.40 & 86.40 & 86.40 & 87.20 & 86.80 & 86.40 & 84.80 \\
Causal judgement & 187 & 48.66 & 48.13 & 55.08 & 50.80 & 54.55 & 53.48 & 44.39 & 50.80 & 53.48 & 50.80 \\
Date understanding & 250 & 32.40 & 35.60 & 40.40 & 39.60 & 44.40 & 49.20 & 43.60 & 38.80 & 40.00 & 50.00 \\
Disambiguation QA & 250 & 50.00 & 48.40 & 50.40 & 52.00 & 50.40 & 47.20 & 53.20 & 51.60 & 58.00 & 46.00 \\
Dyck languages & 250 & 12.80 & 11.60 & 12.80 & 12.40 & 12.00 & 12.00 & 12.40 & 14.80 & 13.20 & 12.40 \\
Formal fallacies & 250 & 14.40 & 25.60 & 54.40 & 55.60 & 41.20 & 46.40 & 50.00 & 48.00 & 49.60 & 46.00 \\
Geometric shapes & 250 & 32.80 & 33.60 & 28.40 & 34.40 & 33.20 & 35.60 & 32.80 & 34.00 & 34.40 & 35.60 \\
Hyperbaton & 250 & 54.80 & 67.60 & 60.80 & 75.20 & 66.40 & 79.60 & 78.40 & 78.00 & 75.20 & 67.60 \\
Logical deduction (5) & 250 & 40.40 & 39.60 & 41.20 & 44.00 & 42.00 & 40.40 & 43.20 & 41.60 & 38.80 & 40.80 \\
Logical deduction (7) & 250 & 23.20 & 24.80 & 23.60 & 24.80 & 26.80 & 29.60 & 27.60 & 22.40 & 26.00 & 26.00 \\
Logical deduction (3) & 250 & 67.60 & 66.40 & 64.80 & 69.60 & 67.20 & 67.20 & 70.40 & 67.60 & 68.40 & 67.60 \\
Movie recommendation & 250 & 40.40 & 16.40 & 50.80 & 56.40 & 41.20 & 57.20 & 49.60 & 40.40 & 47.20 & 50.40 \\
Multistep arithmetic two & 250 & 69.60 & 66.40 & 67.60 & 61.60 & 65.20 & 65.20 & 66.40 & 66.00 & 62.80 & 66.40 \\
Navigate & 250 & 86.40 & 86.40 & 86.80 & 86.80 & 86.40 & 87.60 & 86.00 & 86.00 & 87.60 & 85.60 \\
Object counting & 250 & 78.80 & 76.40 & 78.00 & 78.00 & 77.20 & 76.80 & 79.20 & 80.00 & 79.20 & 79.20 \\
Penguins in a table & 146 & 36.99 & 57.53 & 58.90 & 58.90 & 59.59 & 57.53 & 60.27 & 59.59 & 51.37 & 63.01 \\
Reasoning about colored objects & 250 & 54.80 & 66.40 & 64.00 & 69.60 & 67.60 & 67.20 & 64.40 & 65.60 & 68.80 & 67.60 \\
Ruin names & 250 & 49.60 & 46.80 & 48.80 & 46.80 & 49.20 & 48.40 & 49.60 & 47.60 & 50.80 & 50.40 \\
Salient translation error detection & 250 & 12.40 & 16.80 & 16.00 & 24.80 & 19.20 & 36.00 & 20.00 & 21.20 & 10.40 & 24.00 \\
Snarks & 178 & 6.18 & 7.30 & 8.99 & 11.24 & 10.11 & 12.36 & 8.43 & 8.43 & 7.87 & 12.36 \\
Sports understanding & 250 & 44.40 & 55.20 & 62.40 & 43.60 & 61.60 & 52.00 & 56.80 & 63.60 & 38.00 & 67.60 \\
Temporal sequences & 250 & 32.00 & 31.20 & 27.60 & 29.60 & 26.80 & 26.80 & 28.80 & 26.80 & 29.20 & 31.20 \\
Tracking shuffled objects (5) & 250 & 28.80 & 30.80 & 33.20 & 36.40 & 36.80 & 32.80 & 35.60 & 38.00 & 35.60 & 31.20 \\
Tracking shuffled objects (7) & 250 & 22.00 & 24.40 & 24.80 & 26.80 & 27.20 & 26.40 & 26.80 & 30.00 & 27.20 & 26.40 \\
Tracking shuffled objects (3) & 250 & 45.20 & 44.40 & 48.40 & 56.40 & 54.40 & 44.40 & 57.60 & 53.60 & 56.40 & 46.00 \\
Web of lies & 250 & 100.00 & 100.00 & 100.00 & 100.00 & 100.00 & 100.00 & 100.00 & 100.00 & 100.00 & 100.00 \\
Word sorting & 250 & 8.00 & 9.60 & 12.40 & 11.20 & 12.00 & 9.20 & 12.40 & 11.60 & 11.60 & 10.00 \\
\midrule
Macro average & 6,511 & 43.64 & 45.28 & 48.30 & 49.66 & 48.85 & 49.89 & 49.82 & 49.36 & 48.43 & 49.59 \\
\bottomrule
\end{tabular*}
\setcounter{table}{2}
\captionof{table}{Complete BBH-27 per-task accuracy. $N$ is the number of evaluation examples in each subtask; the total is 6,511. The 3B and IM columns are the original Qwen2.5-3B recipient and Intersection-Merge baseline. T1--T8 are the eight APM slices defined in the main-paper Table~4(a). The final row reports the unweighted macro-average across the 27 tasks and reproduces the averages summarized in Table~4(c). Scores are percentages. All results use an injection ratio of $\mu=0.02$.}
\label{tab:bbh27-complete}
\endgroup
}
\clearpage
\bibliography{aaai2027}
\clearpage
\appendix
\onecolumn
\section*{Supplementary Material}
\addcontentsline{toc}{section}{Supplementary Material}
\BBHTable
\clearpage
\MathSignificanceTable
\vspace{10pt}
\RatioTable
\vspace{8pt}
\ActivationDataTable

\clearpage
\twocolumn
\SupplementaryText
\end{document}